\documentclass[conference]{IEEEtran}
\usepackage{cite}
\ifCLASSINFOpdf
   \usepackage[pdftex]{graphicx}
\else
\fi

\usepackage{amsmath}

\IEEEoverridecommandlockouts

\title{The ParallelEye Dataset: Constructing Large-Scale Artificial Scenes for Traffic Vision Research}

\author{Xuan Li, Kunfeng Wang, \emph{Member, IEEE}, Yonglin Tian, Lan Yan, and Fei-Yue Wang, \emph{Fellow, IEEE}% <-this % stops a space
\thanks{This work was partly supported by National Natural Science Foundation of China under Grant 61533019, Grant 71232006, and Grant 91520301.}% <-this % stops a space
\thanks{Xuan Li is with the School of Automation, Beijing Institute of Technology, Beijing 100081, China, and also with The State Key Laboratory for Management and Control of Complex Systems, Institute of Automation, Chinese Academy of Sciences, Beijing 100190, China (e-mail: lixuan0125@126.com).}%
\thanks{Kunfeng Wang (\emph{Corresponding author}) is with The State Key Laboratory for Management and Control of Complex Systems, Institute of Automation, Chinese Academy of Sciences, Beijing 100190, China, and also with Qingdao Academy of Intelligent Industries, Qingdao 266000, China (e-mail: kunfeng.wang@ia.ac.cn).}%
\thanks{Yonglin Tian is with the Department of Automation, University of Science and Technology of China, Hefei 230027, China, and also with The State Key Laboratory for Management and Control of Complex Systems, Institute of Automation, Chinese Academy of Sciences, Beijing 100190, China.}%
\thanks{Lan Yan is with The State Key Laboratory for Management and Control of Complex Systems, Institute of Automation, Chinese Academy of Sciences, Beijing 100190, China.}%
\thanks{Fei-Yue Wang is with The State Key Laboratory for Management and Control of Complex Systems, Institute of Automation, Chinese Academy of Sciences, Beijing 100190, China, and also with the Research Center for Computational Experiments and Parallel Systems Technology, National University of Defense Technology, Changsha 410073, China (e-mail: feiyue@gmail.com).}%
}

\begin{document}
\maketitle


\begin{abstract}
Video image datasets are playing an essential role in design and
evaluation of traffic vision algorithms. Nevertheless, a
longstanding inconvenience concerning image datasets is that
manually collecting and annotating large-scale diversified datasets
from real scenes is time-consuming and prone to error. For that
virtual datasets have begun to function as a proxy of real datasets.
In this paper, we propose to construct large-scale artificial scenes
for traffic vision research and generate a new virtual dataset
called ``ParallelEye''. First of all, the street map data is used to
build 3D scene model of Zhongguancun Area, Beijing. Then, the
computer graphics, virtual reality, and rule modeling technologies
are utilized to synthesize large-scale, realistic virtual urban
traffic scenes, in which the fidelity and geography match the real
world well. Furthermore, the Unity3D platform is used to render the
artificial scenes and generate accurate ground-truth labels, e.g.,
semantic/instance segmentation, object bounding box, object
tracking, optical flow, and depth. The environmental conditions in
artificial scenes can be controlled completely. As a result, we
present a viable implementation pipeline for constructing
large-scale artificial scenes for traffic vision research. The
experimental results demonstrate that this pipeline is able to
generate photorealistic virtual datasets with low modeling time and
high accuracy labeling.
\end{abstract}

% \fi
\IEEEpeerreviewmaketitle
\section{Introduction}
The publicly available video image datasets have received much attention in recent years, due to its
indispensability in design and evaluation of computer vision algorithms \cite{Geiger2013}.
In general, a computer vision algorithm needs a large amount  of  labeled images for
training and evaluation. The datasets can be divided into two types:
unlabeled datasets used for unsupervised learning and labeled datasets used for supervised
learning.
However, manually annotating the images is
time-consuming and labor-intensive, and participants often lack professional
knowledge, making some annotation tasks difficult to execute.
Experts are always sparse  and should be properly
identified.
As we known, the human annotators are subjective, and their annotations should be re-examined if two or more annotators have disagreements about the label of one entity. By contrast, the computer is objective in processing data and particularly good at batch processing,
so why not let the computer annotate the images automatically?

At present, most publicly available datasets are obtained from real scenes.
As the computer vision field enters the big data era, researchers begin to
look for better ways to annotate large-scale datasets \cite{Handa2014}. At the same
time, the development of virtual datasets has a long history,
starting at least from Bainbridge's work \cite{Bainbridge2007}. Bainbridge used
Second Life and World of Warcraft as two distinct examples of
virtual worlds to predict the scientific research potential of virtual worlds, and introduced the virtual worlds into a
lot of research fields that scientists are now exploring,
including sociology, computer science, and
anthropology.
In fact, synthetic data has been used for decades to
benchmark the performance of computer vision algorithms. The use of
synthetic data has been particularly significant in object detection
[4], [5] and optical flow estimation [6]-[8], but most virtual
data are not photorealistic or akin to the real-world data, and lack
sufficient diversity \cite{Ros2015}. The fidelity of some virtual data is
close to the real-world \cite{Prendinger2013}. However, the synthesized virtual worlds are
seldom equivalent to the real world in geographic
position, and seldom annotate the virtual images automatically. Richter \emph{et al.} \cite{Richter2016}
used a commercial game engine to extract virtual images, with no access
to the source code or the content. The SYNTHIA dataset \cite{Ros2016}
provided a realistic virtual city as well as
synthetic images with automatically generated pixel-level annotations, but in that dataset there lacks
other annotation
information such as object bounding box and object tracking. Gaidon \emph{et al.} \cite{Gaidon2016} proposed a virtual dataset called ``Virtual KITTI" as a proxy for tracking algorithm evaluation. While this dataset was cloned from ``KITTI", it cannot extend easily to arbitrary traffic networks.
Due to the above limitations, new virtual datasets that match the real world and provide detailed ground truth annotations are still desirable.

\begin{figure*}[!t]
\centering
\includegraphics[width=7in]{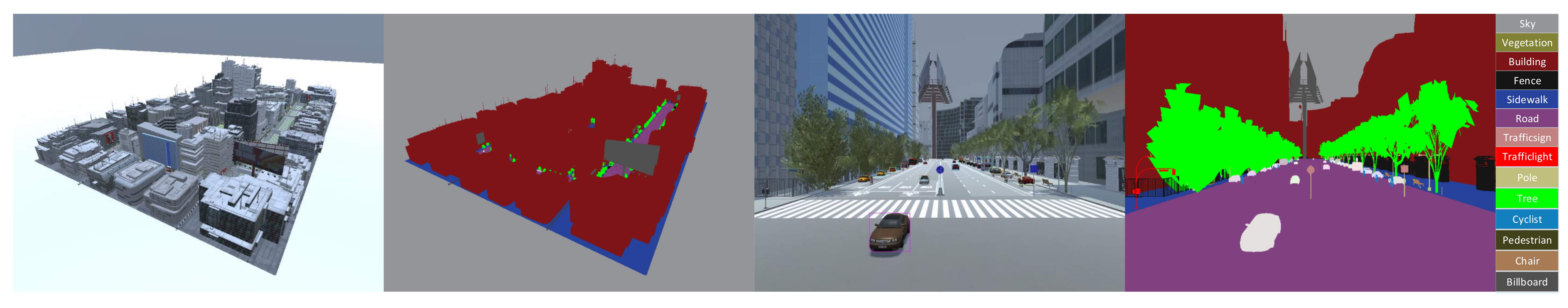}
\caption{Examples of our generated ParallelEye dataset. From  left to right: a
general view of the constructed artificial scenes, its semantic labels, a sample
frame with tracking bounding boxes generated automatically, and its
semantic labels. Best viewed with zooming.} \label{fig_sim}
\end{figure*}

Manually annotating pixel-level semantics for images is time-consuming and not accurate enough.
For example, annotating
high-quality semantics with 10-20 categories in one image usually takes 30-60
minutes \cite{Kundu2014}. This is known as the ``curse of dataset annotation'' \cite{Xie2016}.
The more detailed the semantics,
the more labor-intensive the annotation process.
As a result, many datasets do not provide semantic
segmentation annotations. For example, ImageNet \cite{Karpathy2014},\cite{Russakovsky2015} has 14 million images, in which
more than one million images have definite class and the images are
annotated with object bounding box for object recognition. However,
ImageNet does not have semantic segmentation annotations.
Some datasets provide only limited semantic segmentation
annotations. For example, NYU-Depth V2 \cite{Silberman2012} has 1449
densely labelled images, KITTI \cite{Geiger2013} has 547 images, CamVid \cite{Brostow2009},\cite{Browstow2008} has 600
images, Urban LabelMe \cite{Russell2008} has 942 images, and Microsoft COCO \cite{Lin2014} has
three hundred thousand images. These datasets play an important role
in the study of semantic segmentation. However, these datasets cannot
be used directly in intelligent transportation, especially in automobile navigation,
because the number of labeled images is insufficient and the segmented semantics
have different categories. Currently, computer vision algorithms
that exploit context for pattern recognition would benefit from
datasets with many annotated categories embedded in images from complex
scenes. Such datasets should contain a wide variety of environmental
conditions with annotated object instances co-occurring in the same scenes.
However, the real scenes are unrepeatable and the captured images are expensive to annotate,
making it difficult to obtain large-scale, diversified datasets with precise annotations.

In order to solve these problems, this paper proposes a pipeline for constructing artificial scenes and generating virtual images.
First of all, we use map data to build the 3D scene model of Zhongguancun Area, Beijing.
Then, we use the computer graphics, virtual reality, and rule
modeling technologies to create a realistic, large-scale virtual urban
traffic scene, in which the fidelity and geographic
information can match the real world well. Furthermore, we use the Unity3D development platform
for rendering the scene and automatically annotating the ground truth labels including  pixel-level semantic/instance segmentation, object bounding box, object tracking, optical flow,
and depth. The environmental
conditions in artificial scenes can be controlled completely. In consequence, we generate a new virtual image dataset, called ``ParallelEye" (see Fig. 1). We will build a website and make this dataset publicly available before the publication of this paper. The
experimental results demonstrate that our proposed implementation pipeline is able to
generate photorealistic virtual images with low modeling time
and high fidelity.

\begin{figure}[!t]
\centering
\includegraphics[width=3.3in]{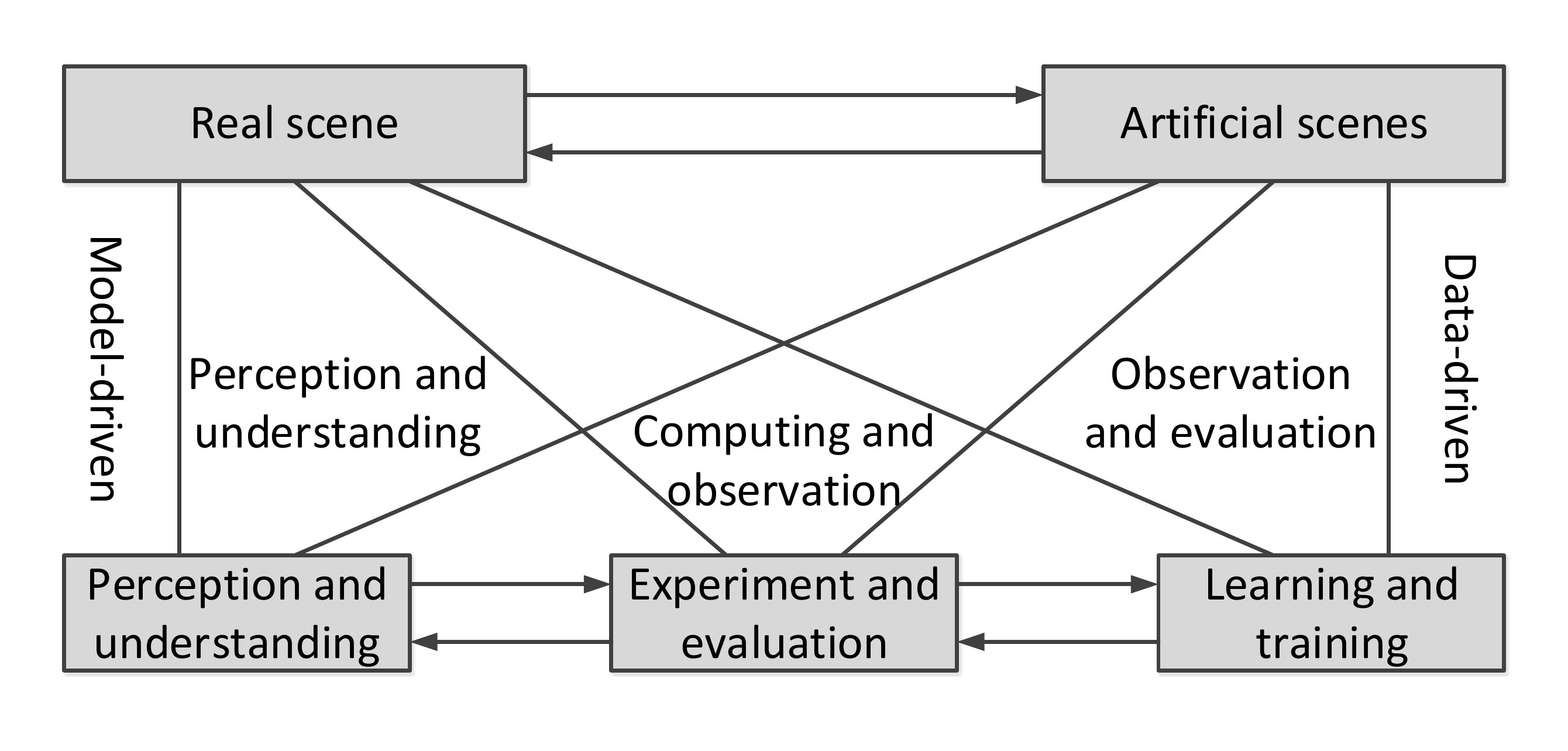}
\caption{Basic framework and architecture for parallel vision \cite{KWang2016}.}
\label{fig_sim}
\end{figure}

The rest of this paper is organized as follows. Section II
introduces the significance of parallel vision and virtual dataset. Section III
presents our approach to constructing artificial scenes and generating virtual images with
ground-truth labels. Section IV reports the experimental
results and analyzes the performance. Finally, the concluding remarks are made in section V.


\section{Parallel Vision and Virtual Dataset}

Parallel vision [23]-[25] is an extension of the ACP
(Artificial systems, Computational experiments, and Parallel
execution) theory [26]-[30] into the computer vision field. For parallel vision,
photo-realistic artificial scenes are
used to model and represent complex real scenes, computational experiments are utilized to learn and evaluate a variety of
vision models, and parallel execution is conducted to online optimize the vision system and realize perception and understanding
of complex scenes.
The basic framework
and architecture for parallel vision \cite{KWang2016} is shown in Fig. 2.
Based on the parallel vision theory,
this paper constructs a large-scale virtual urban network and synthesizes a large number of
realistic images.

The first stage of parallel vision is to construct photorealistic artificial scenes by simulating a variety of environmental conditions occurring in real scenes, and accordingly to synthesize large-scale diversified datasets with precise annotations generated automatically. Generally speaking, the construction of artificial scenes can be regarded as ``video game design", i.e., using the computer animation-like techniques to model the artificial scenes. The main technologies used in this stage include computer graphics, virtual reality, and micro-simulation. Computer graphics and computer vision, on the whole, can be thought of as a pair of forward and inverse problems. The goal of computer graphics is to synthesize image measurements given the description of world parameters according to physics-based image formation principles (forward inference), while the focus of computer vision is to map the pixel measurements to 3D scene parameters and semantics (inverse inference). Apparently their goals are opposite, but can converge to a common point: parallel vision.

%In many situations, due to the difficulty in data collection and annotation, we are unable to obtain a satisfying training dataset from real scenes. This situation is bound to hinder the design and evaluation of computer vision algorithms. Fortunately, the synthetic dataset compiled from artificial scenes can function as a proxy of real dataset. Firstly, by means of off-the-shelf computing resources, artificial scenes are ready to generate an infinite amount of training data, and by flexibly configuring the ingredients (e.g., scene layout, illumination, weather, and camera viewpoint) of artificial scenes, we can synthesize images with sufficient diversity. In consequence, it becomes possible to meet the requirement for large-scale diversified datasets. Secondly, while the real scene is usually unrepeatable, computer-generated artificial scenes can be repeated easily. By fixing some physical models and parameters while tuning the others, we can customize the image formation factors for artificial scenes. This allows us to evaluate computer vision algorithms from various angles. Last but not least, there are some particular real scenes from which it is impossible to capture valuable training images, while the artificial scenes are seldom limited. To sum up, constructing artificial scenes is of great significance, given that it offers a reliable data source (as supplement to real-scene data) for design and evaluation of computer vision systems.

From the parallel vision perspective, we
design the ParallelEye dataset.
ParallelEye is synthesized by referring to the urban network of Zhongguancun Area, Beijing.
Using OpenStreetMap (OSM), an urban network with length 3km and width 2km is extracted.
Artificial scenes are constructed on this urban network.
Unity3D is used to control the environmental conditions in the scene.
There are
15 object classes in ParallelEye, reflecting the common elements of
traffic scenes, including sky, buildings, cars, roads, sidewalks,
vegetation, fence, traffic signs, traffic lights, lamp poles,
billboards, trees, cyclists, pedestrians, and chairs.
These object classes
can be automatically annotated to generate pixel-level
semantics.
For traffic vision research, we pay special attention to instance segmentation, with each object of interest segmented automatically.
In addition, ParallelEye
provides accurate ground truth for object
detection and tracking, depth, and optical flow.

\section{Approach}
Our pipeline for generating the ParallelEye dataset is shown in Fig. 3. Firstly,
the OSM data released by OpenStreetMap is used to achieve the
correspondence in geographic location between the virtual and real world.
Secondly, CityEngine is used to write CGA (Computer Generated Architecture) rules and
design a realistic artificial scene, including roads, buildings, cars, trees, sidewalks, etc.
Thirdly, the artificial scene is imported into Unity3D
 and gets rendered by using the script and the shader. In the
dataset, accurate ground truth annotations are generated automatically,
and environmental conditions can be controlled completely and flexibly.

\begin{figure}[!t]
\centering
\includegraphics[width=2.35in]{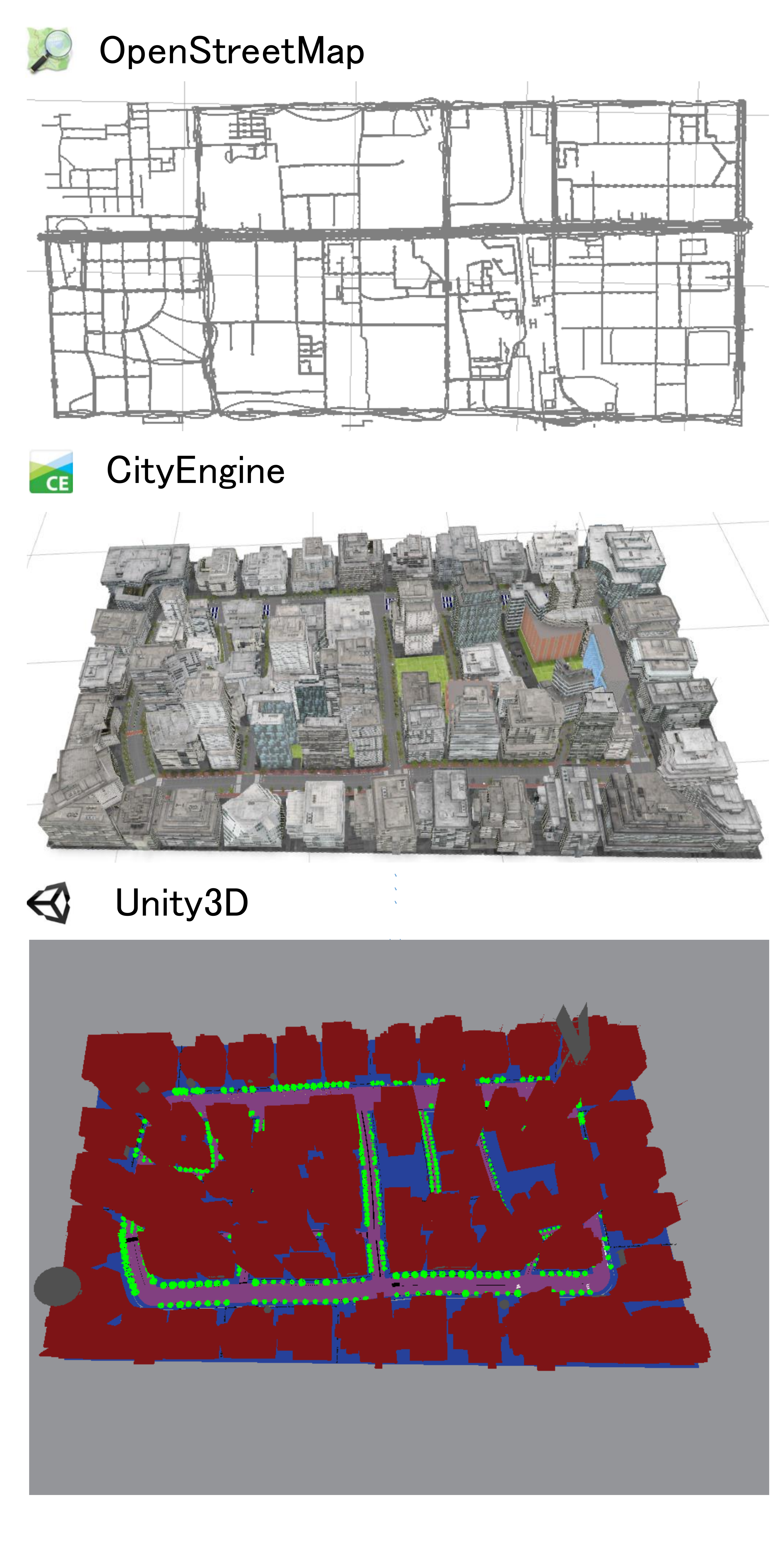}
\caption{Pipeline for generating the ParallelEye dataset with OpenStreetMap, CityEngine,
and Unity3D.} \label{fig_sim}
\end{figure}

\subsection{Correspondence of Artificial and Real Scenes}

In order to increase the
fidelity, we choose to import geographic data from OpenStreetMap. Although Google Maps occupies an important
position in geographic information, it is not an open-source
software. By contrast, OpenStreetMap is an open-source, online map editing program with the
goal of creating a world where content is freely accessible to
everyone.
 In OpenStreetMap, the ways denote a
directional node sequence. Each node of the network can connect 2-2000 paths, and then arrive at another node.
The road information includes direction, lane number, lane
width, street name, and speed limit. Each path can form
three combinations: non-closed paths, closed paths, and regions. The
non-closed paths correspond to the roads, rivers, and railways in
the real world. The closed paths correspond to subway, bus routes,
residential roads, and so on. The regions correspond to buildings,
parks, lakes, and so on. Based on the properties of OSM data, it
is easy to relate the real world to the geographic information of
the artificial scene. Fig. 4 shows the real Automation Building of CASIA (Institute of Automation, Chinese Academy of
Sciences) and its virtual proxy generated by CGA rules. They are similar in appearance.

\begin{figure}[!t]
\centering
\includegraphics[width=2.35in]{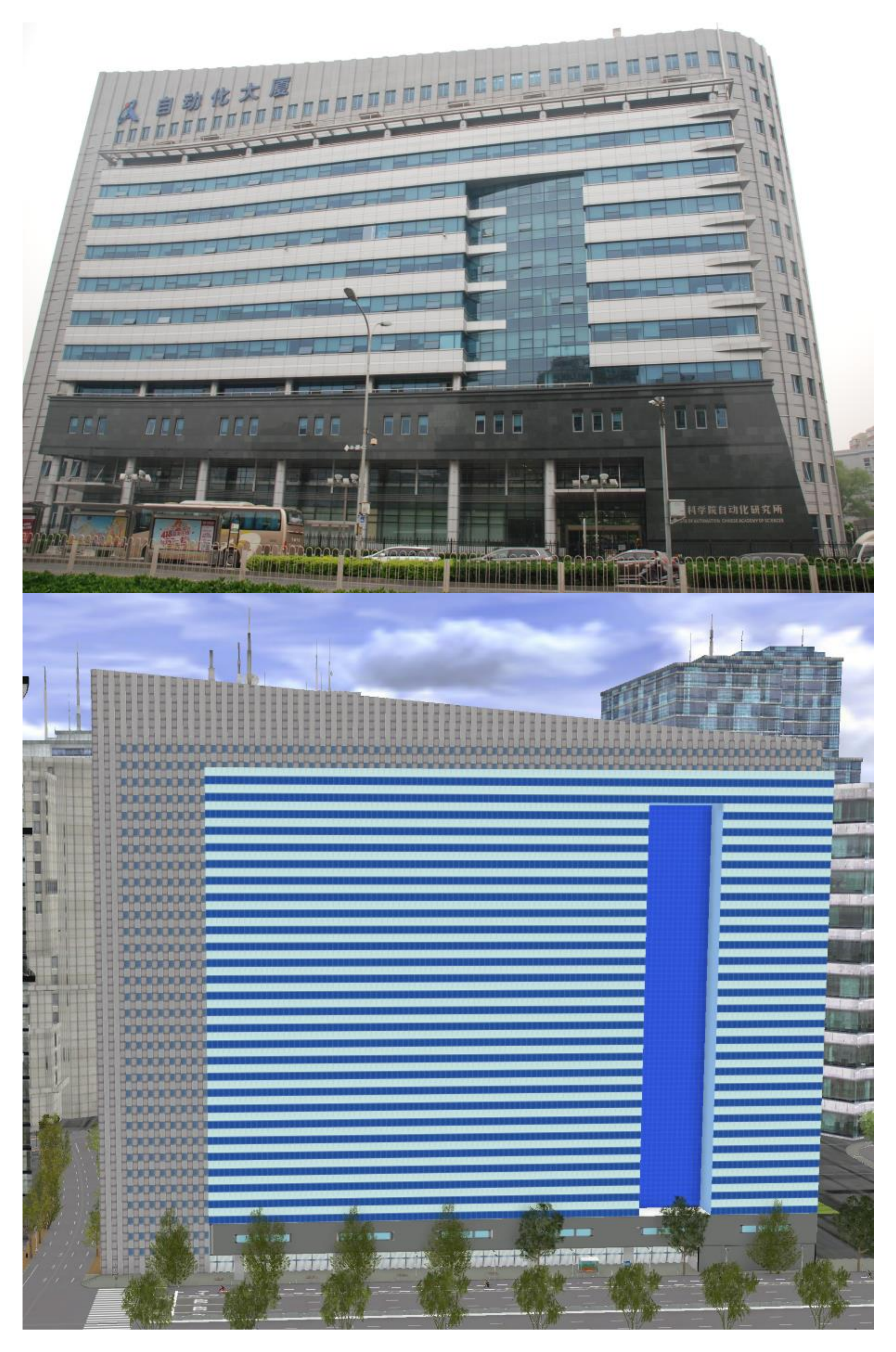}
\caption{The real Automation Building of CASIA (top) and its virtual proxy (bottom).}
\label{fig_sim}
\end{figure}

\subsection{Generation of Ground-Truth Annotations}
As stated above, ground-truth annotations are essential for
vision algorithm design and evaluation. Traditionally, the images were annotated by hand.
The manual annotation is time-consuming and prone to error. Taking
semantic/instance segmentation as an example, it usually takes 30-60 minutes to annotate an image with 10-20
object categories.
Besides, manual annotation is more or less subjective, so that different annotators can make different semantic labels for the same image, especially near the object boundaries.
Instead of manual annotation, this paper
uses Unity3D to automatically generate
accurate ground-truth labels.
Fig. 5 shows some examples of ground-truth annotations, including depth, optical flow, object tracking, object detection, instance segmentation, and semantic segmentation.

Generating ground truth with Unity3D is accurate and efficient.
Semantic
segmentation ground truth can be directly generated by
using unlit shaders on the materials of the objects, with each category
outputting a unique color.
Instance segmentation
ground truth is generated using the same method, but assigns a unique color tag to
each object of interest.
The modified shaders output a color
which is not affected by the lighting and shading conditions.
Depth ground truth is generated using built-in depth buffer information to get
depth data for screen coordinates. The depth ranges from 0 to
1 with a nonlinear distribution, with 1 representing ``infinitely distant". Optical flow
ground truth is generated by calculating the instantaneous velocity of moving objects on the imaging plane and using the pixel changes in the image sequence to find the
correspondence between the previous frame and the current frame. Given a
pixel point $(x,y)$ in the image, at any time the brightness of that point
is $E(x+\triangle x,y+\triangle y,t+\triangle t)$.
Let  $(u,v)=(\frac{\partial x}{\partial
t},\frac{\partial y}{\partial t})$ represent the instantaneous velocity of the point in the
horizontal and vertical directions, the brightness change occurs when
the point moves. We use the Taylor formula to represent the pixel brightness:
\begin{equation}\label{}
\begin{split}
& E(x+\triangle x,y+\triangle y,t+\triangle t)  \\
& =E(x,y,t)+\frac{\partial E}{\partial x}\triangle x+\frac{\partial
E}{\partial y}\triangle y+\frac{\partial E}{\partial t}\triangle
t+\varepsilon.
\end{split}
\end{equation}
For any $\triangle t\rightarrow0$, let $\omega=(u,v)$ , the optical flow constraint
equation is given by
\begin{equation}\label{}
-\frac{\partial E}{\partial t}=\frac{\partial E}{\partial
x}\frac{\partial x}{\partial t}+\frac{\partial E}{\partial
y}\frac{\partial y}{\partial t}=\nabla E \cdot \omega,
\end{equation}
where $\omega$ is the optical flow of $E(x,y,t)$.

We generate multi-object tracking ground truth based on
 four rules: 1) when the object appears
within the field of view of the camera, the three-dimensional bounding box
of the object is converted to a two-dimensional bounding box; 2)
when the object appears or disappears from the image
boundary, we perform special handling for the bounding box; 3)
we do not draw bounding boxes for objects that have less than 15 pixels in width or less than 10
pixels in height; 4) when occlusion
occurs and the occlusion rate is higher than a
threshold, we do not draw bounding boxes for the occluded object.

\begin{figure}[!t]
\centering
\includegraphics[width=3.5in]{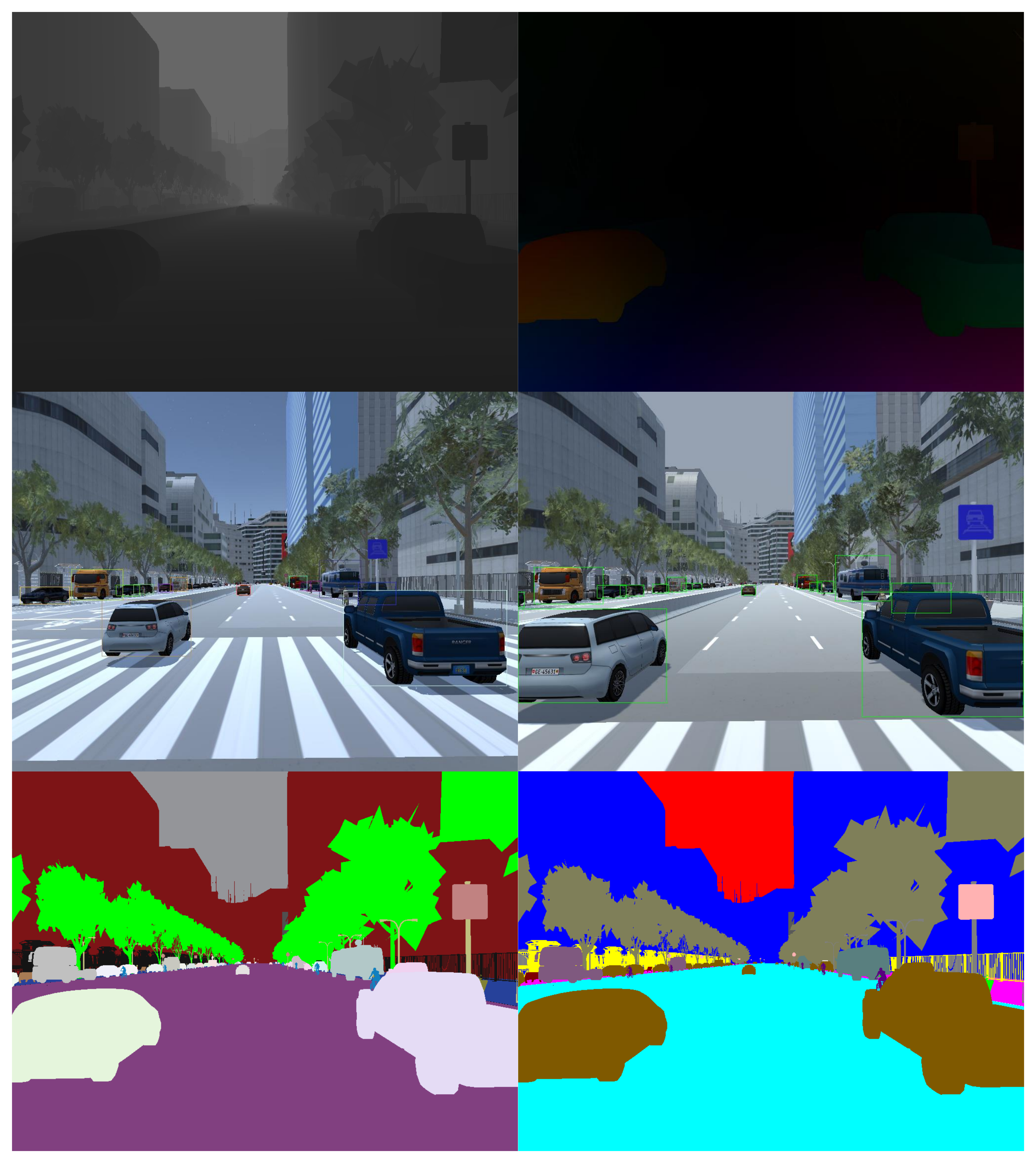}
\caption{Examples of ground-truth annotations generated  automatically by Unity3D. Top: depth (left) and optical flow (right). Middle: object tracking (left) and object detection (right). Bottom: pixel-level instance segmentation (left) and semantic
segmentation (right). Best viewed with zooming.} \label{fig_sim}
\end{figure}

\subsection{Diversity of Artificial Scenes}
In order to increase the diversity and fidelity of artificial scenes, we control the parameters in the script, the material, and the simulated environmental conditions. Specifically, the controllable parameters include: 1)
number, type, trajectory, speed, and direction of
the vehicles; 2) position and configuration of the camera; 3) weather (sunny, cloudy,
rainy, foggy, etc) and illumination (daytime, dawn, dusk, etc).

\begin{figure}[!t]
\centering
\includegraphics[width=3.5in]{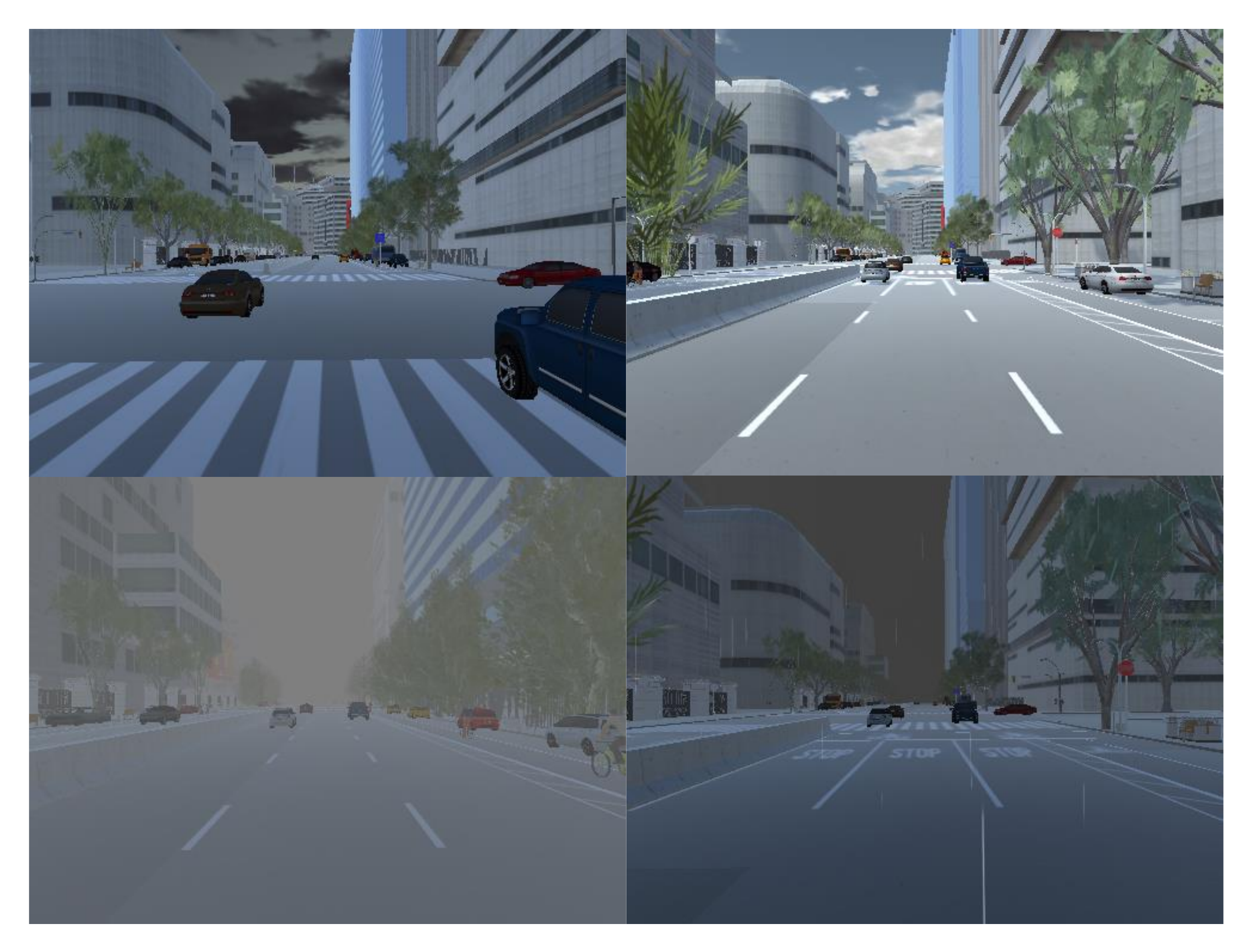}
\caption{Illustration of the diversity of artificial scenes. Top: Virtual images with illumination at 6:00 am (left) and
12:00 pm (right) in a sunny day. Bottom: Virtual images with weather of fog (left) and
rain (right).} \label{fig_sim}
\end{figure}

Traditionally, video image datasets are collected by capturing in the real world or retrieving from the Internet. It is impossible to control the environmental conditions and repeat the scene layout under different environments, and thus difficult to isolate the effects of
environmental conditions on the performance of computer vision algorithms.
By contrast, it is
easy to control the environmental conditions
in artificial scenes.
In this work, we are able to flexibly control
the camera's location, height, and orientation to
capture different contents of the
artificial scene.
We are also able to dynamically change the illumination (from sunrise to sunset) and weather conditions (sunny, cloudy, and foggy). Although
we can change  the environmental conditions in artificial scene, the ground-truth annotations are always easy to generate,
no matter how adverse the illumination and weather conditions are and how
blurred the image details are. This makes it possible to quantitatively analyze the impacts of each environmental condition on algorithm performance, usually called ``ceteris
paribus analysis". Fig. 6 illustrates the diversity of artificial scenes in terms of illumination and weather conditions.

\section{Experiments}
Based on the proposed approach, we construct the artificial scene and configure virtual cameras to capture images from the scene. The virtual cameras can be moving or stationary. For automobile applications, the virtual cameras are installed on moving vehicles. For visual surveillance applications, the virtual cameras are fixed on the roadside or at intersections. The experiments are conducted to verify that the artificial scenes are repeatable and that the camera's position, height, and orientation can be configured flexibly.

\subsection{Onboard Camera}
In this experiment, an onboard camera is configured at a height of 2 meters, mimicking
the camera installed on the vehicle roof. There are totally 67 vehicles on the road,
including 52 vehicles parking on the roadside (3 buses, 4 trucks,
and 45 cars ) and another 15 vehicles in motion.
We turn the camera orientation
from left to right and get five orientations (i.e., -30, -15, 0, 15, and 30 degrees with respect to the lane direction). The distance between two cameras on adjacent lanes is 5 meters. These configurations lead to
substantial changes in object appearance. Fig. 7 shows three continuous images captured by the onboard
camera.

\begin{figure}[!t]
\centering
\includegraphics[width=3.5in]{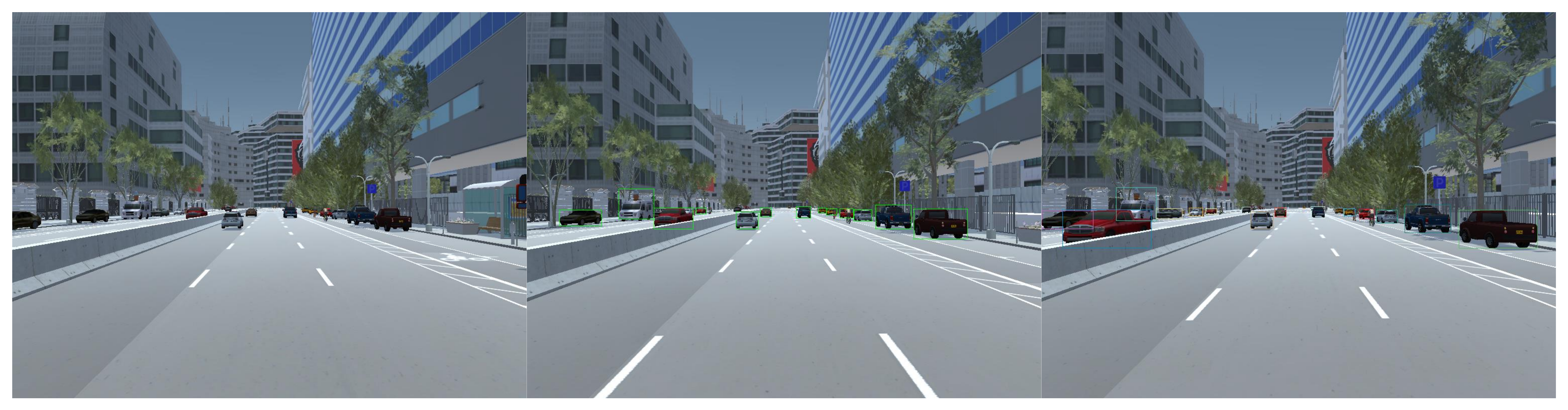}
\caption{Continuous images captured by an onboard
camera: a sample image (left), another image annotated with object bounding boxes (middle), and the third image annotated with tracking bounding boxes of different colors (right). Best viewed with zooming.} \label{fig_sim}
\end{figure}

\subsection{Surveillance Camera}

In this experiment, a surveillance camera is installed at an intersection. We rotate the camera and control the rotation speed at 10 degrees per second, and the
rotation range is 180 degrees.
We also change the camera height, with the lifting speed of 0.1 meters per second and the lifting range of 2-5 meters.
Such settings can fully simulate the role of surveillance cameras.
Based on this experiment,
the artificial scene provides virtual video images for intersection monitoring.
Fig. 8 shows images captured by the surveillance
camera.

\begin{figure}[!t]
\centering
\includegraphics[width=3.5in]{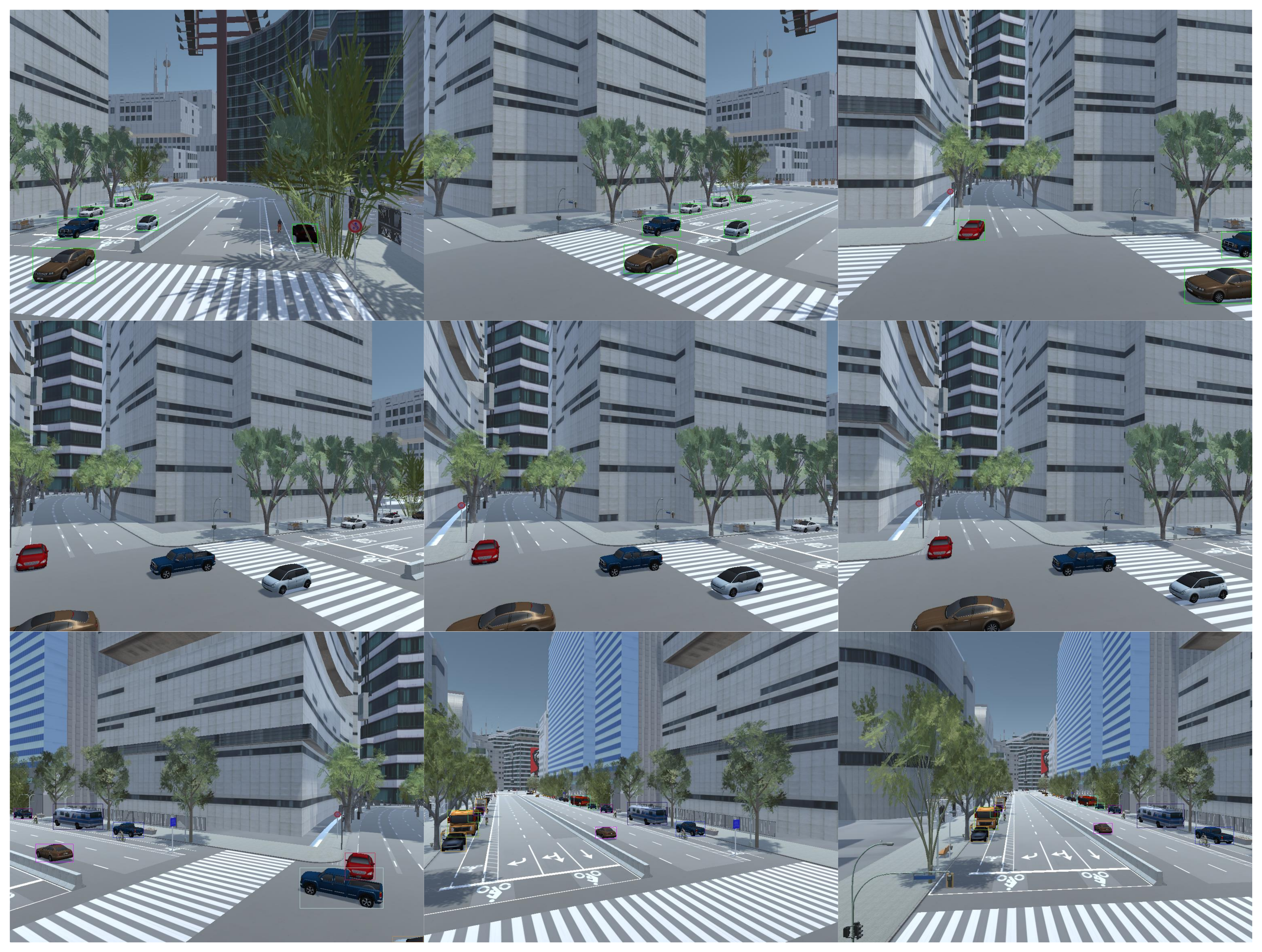}
\caption{Continuous images captured by a surveillance
camera: images annotated with object bounding boxes (top row), original images (middle row), and images
annotated with tracking bounding boxes of different colors (bottom row). Best viewed with zooming.} \label{fig_sim}
\end{figure}

In order to increase diversity of virtual images and record the ground truth, we adopt the same operations for both the onboard camera and the surveillance camera. To record the ground truth, we use a green bounding box to record the detection ground truth for each object. We also assign a bound box of unique color to record the tracking ground truth for each object instance. To increase diversity, we dynamically change the illumination (daytime, dawn, and dusk) and weather (sunny, cloudy, rainy, and foggy) conditions in the artificial scenes. These
subtle changes simulate different environmental conditions in the virtual world, and
would otherwise need the expensive process of re-acquiring and
re-labeling images of the real world. The advantage of this setting is
that it can increase diversity of the ParallelEye
dataset.
In the experiments, with image resolution of 500*375 pixels for ParallelEye, the pipeline for artificial scene construction and ground truth generation runs at
8-12 fps (frames per second) on a workstation computer. We have collected a total of 31,000 image frames, each of which has been annotated with accurate ground truth. We will build a website and make the dataset publicly available before the
publication of this paper.

\section{Concluding Remarks}

In this paper, we propose a new virtual image
dataset called ``ParallelEye". For that we present a dataset
generation pipeline that uses street map, computer graphics, virtual reality, and
rule modeling technologies to construct a realistic, large-scale virtual
urban traffic scene. The artificial scene matches the real world well in terms of fidelity and
geographic information.
In the artificial
scene, we flexibly configure the camera (including its position, height,
and orientation) and the environmental conditions, to collect diversified images. Each image has been annotated automatically with ground truth including semantic/instance segmentation, object bounding
box, object tracking, optical flow, and depth.

In the future, we will improve the diversity of ParallelEye by
introducing moving pedestrians and cyclists, which are
harder to animate. We will increase the scale of ParallelEye.
In addition, we will combine ParallelEye and the existing real
datasets (e.g., PASCAL VOC, MS COCO, and KITTI) to learn and evaluate
traffic vision models, in order to improve the accuracy and robustness of
traffic vision models when applied to complex traffic scenes.

\end{document}